\documentclass[letterpaper, 10pt, conference]{ieeeconf}         

\IEEEoverridecommandlockouts                              
\usepackage{multirow}
\usepackage{comment}
\usepackage{graphicx}
\usepackage{fixltx2e}
\usepackage{hyperref}
\usepackage{mathtools}
\usepackage{amssymb}
\usepackage{float}
\usepackage{subcaption}

\usepackage[labelsep=endash]{caption}

\title{\LARGE \bf
Human Activity Recognition in RGB-D Videos by Dynamic Images}

\author{Snehasis Mukherjee, Leburu Anvitha and T. Mohana Lahari 
}

\begin{document}

\maketitle
\thispagestyle{empty}
\pagestyle{empty}

\begin{abstract}
Human Activity Recognition in RGB-D videos has been an active research topic during the last decade. However, no efforts have been found in the literature, for recognizing human activity in RGB-D videos where several performers are performing simultaneously. In this paper we introduce such a challenging dataset with several performers performing the activities. We present a novel method for recognizing human activities in such videos. The proposed method aims in capturing the motion information of the whole video by producing a dynamic image corresponding to the input video. We use two parallel ResNext-101 to produce the dynamic images for the RGB video and depth video separately. The dynamic images contain only the motion information and hence, the unnecessary background information are eliminated. We send the two dynamic images extracted from the RGB and Depth videos respectively, through a fully connected layer of neural networks. The proposed dynamic image reduces the complexity of the recognition process by extracting a sparse matrix from a video. However, the proposed system maintains the required motion information for recognizing the activity. The proposed method has been tested on the MSR Action 3D dataset and has shown comparable performances with respect to the state-of-the-art. We also apply the proposed method on our own dataset, where the proposed method outperforms the state-of-the-art approaches.
\end{abstract}
RGB-D video, Human Activity Recognition, Dynamic images, ResNext-101, Depth map.

\section{Introduction}

Human action recognition in videos, has been an active area of interest for the researchers of computer vision and robotics. Automatic recognition of human activities at secured places like airport, rail station, bus stops, shopping areas is necessary for several purposes including security, surveillance, elderly and child monitoring and robotics. After the introduction of depth based cameras such as Microsoft kinect, the researchers working in the area of human activity recognition, have got a handful of sophisticated tools to get the depth and skeleton information. Hence, sophisticated techniques are necessary to make use of the available depth information to recognize human activities accurately. Recently several efforts are being made for recognizing human activity in RGB-D videos \cite{mtap_survey}.
\begin{figure}
\centering{
\includegraphics[width=4.2cm,height=3.6cm]{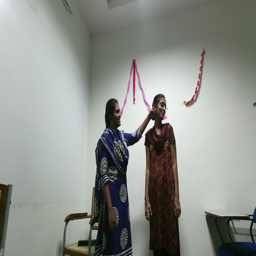}
\includegraphics[width=4.2cm,height=3.6cm]{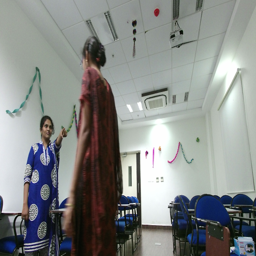}}
\centerline{(a)\hspace{3cm}(b)}
\centering{
\includegraphics[width=4.2cm,height=3.6cm]{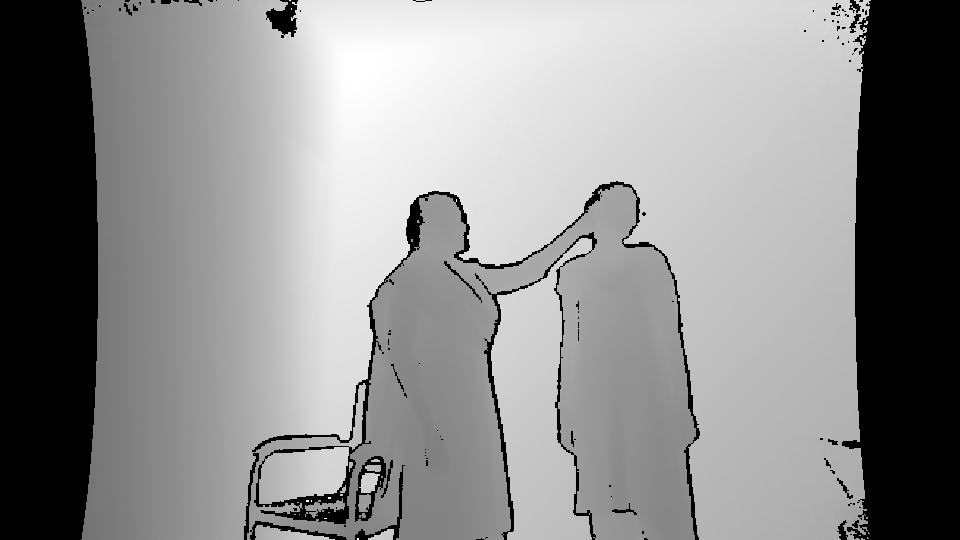}
\includegraphics[width=4.2cm,height=3.6cm]{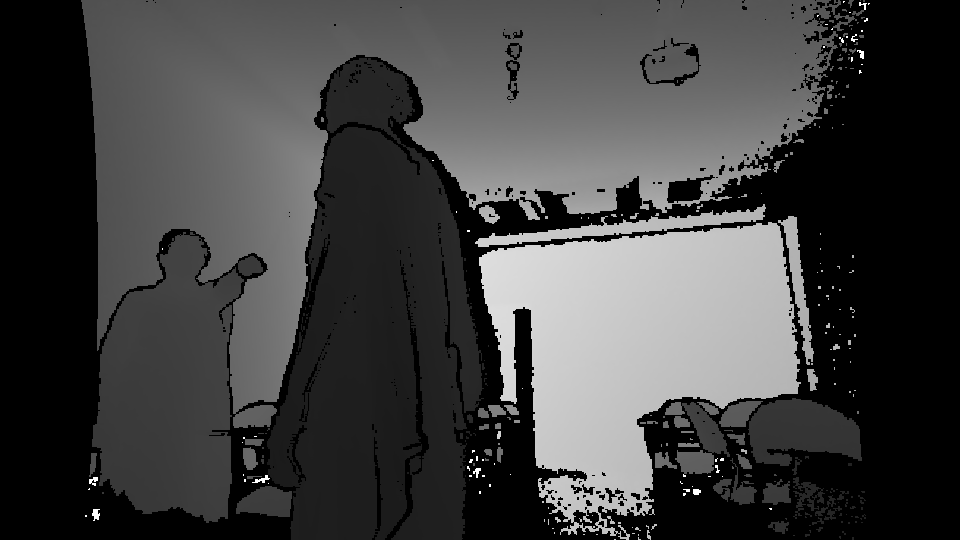}}
\centerline{(c)\hspace{3cm}(d)}
\caption{Example frames from sample videos from ``Pointing'' and ``Punching'' activities of the proposed dataset. (a) RGB frame of ``Punching'', (b) RGB frame of ``Pointing'', (c) Depth frame of ``Punching'', (d) Depth frame of ``Pointing''. The RGB frames may confuse between the 2 activities, whereas, the depth frames can clearly distinguish between the activities.}
\label{fig1}
\end{figure}

Most of the depth based human activity recognition approaches rely on spatial and temporal motion information of the video to extract features, and extend the features for depth frames  \cite{mtap_survey}. However, RGB based features often may not be able to utilize the depth information, necessary for recognizing the activity. For example, when multiple performers are performing simultaneously in a video, the spatio-temporal features are often found to be unable to capture the inherent depth information of the RGB-D video, to classify similar actions. The main challenge in human activity recognition in RGB-D videos, lies in efficiently extracting the motion information in three directions (the x-, y- and z-directions), which still remains an unexplored research area.

After introduction of the deep learning based methods, researchers have made several attempts to replace the hand-crafted features by efficient deep learned features extracted using different well-known neural network models \cite{deep_survey}. However, analyzing the motion using depth information, remains a less-explored area of research for depth-based activity recognition.

Several datasets have been proposed to validate methods for human activity recognition in RGB-D videos \cite{data_survey}. With the advancements of deep learning techniques, a much bigger dataset became necessary to validate a deeper network. Shahroudy \textit{et al.} proposed a sufficiently large dataset with 60 different action classes and more than 56 thousand videos \cite{ntu_dataset}. However, the RGB-D human activity recognition datasets available in the literature, lack in representing activities which are ambiguous in their appearance. For example, ``Pointing'' and ``Punching'' activities may appear as the same, without a depth information, as shown in Figure \ref{fig1}.

\begin{figure}
\centering{
\includegraphics[width=4.2cm,height=3.6cm]{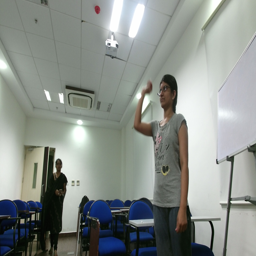}
\includegraphics[width=4.2cm,height=3.6cm]{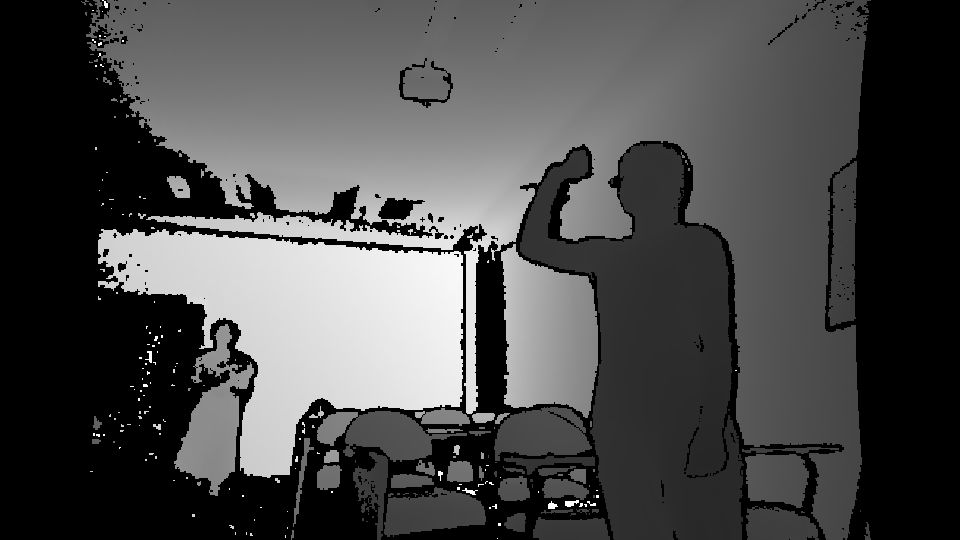}}
\centerline{(a)\hspace{3cm}(b)}
\centering{
\includegraphics[width=4.2cm,height=3.6cm]{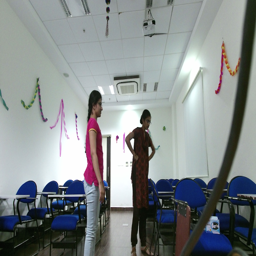}
\includegraphics[width=4.2cm,height=3.6cm]{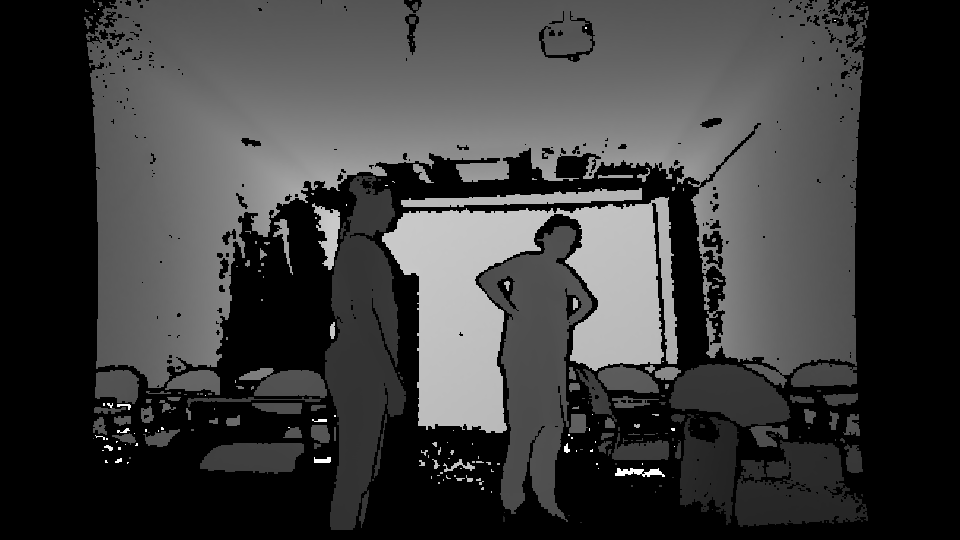}}
\centerline{(c)\hspace{3cm}(d)}
\caption{Example frames from sample videos from ``High Throw'' and ``Pick \& Throw'' activities of the proposed dataset. (a) RGB frame of ``High Throw'', (b) Depth frame, (c) RGB frame of ``Pick \& Throw'', (d) Depth frame. In both the cases one more person is visible other than the performers, which may confuse the recognition process.}
\label{fig2}
\end{figure}

In this study, we propose a challenging RGB-D dataset, consisting of 7 activity classes. The proposed dataset is not a huge in size, but offers two major challenges which none of the existing datasets do. First, the proposed dataset consists of several ambiguous action classes (such as ``Pointing'' and ``Punching'', which are difficult to distinguish without the use of depth information. Second, the proposed dataset contains several videos where several persons are performing some activities simultaneously, at the background. In such scenario, finding the key performers in the video and eliminate the effect of motion information of the background is the key challenge in the proposed dataset. Figure \ref{fig2} shows examples of some frames taken from some videos of the proposed dataset, where multiple persons are performing some activities.

In order to address the challenges of handling the motion information of multiple actors, we propose a technique based on dynamic images extracted from the RGB and Depth videos separately, following \cite{c18}. The concept of dynamic image, introduced by Bilen \textit{et al.}, is a compact representation of a video, encoding the temporal data such as optical flow by rank pooling, particularly in combination with convolutional neural networks (CNNs). Precisely, dynamic image summarizes the video dynamics maintaining the appearance of moving objects (such as human, during an activity). The main advantage of using dynamic images converting any video into an image is that, the existing simple 2D CNN models pre-trained with still images can be applied on the videos.

We apply ResNext-101 \cite{resnext} on the dynamic images extracted from RGB and depth videos separately after a processing step on the dynamic images. During processing step, we apply a probabilistic model based on the Gestalt philosophy following \cite{mtap}, to extract the motion information of the performers during the activity, and minimize all the unnecessary motion information (e.g., background motion, other moving persons and objects). We supply the output vectors of ResNext-101 applied on the RGB and Depth videos to a fully connected layer to come up with a descriptor combining the motion information from the RGB and depth videos.

In this paper, our contributions are three-folds. First, we have introduced a challenging RGB-D activity Recognition dataset. Second, we have performed the activity recognition task on dynamic images, which reduces lots of memory overhead during execution, which is a very common problem while working on RGB-D videos, due to the depth frames. Third, we minimize the effect of background motion information of the video, by applying a threshold on the depth dynamic image intensity values, based on Gestalt theory.

\section{Related Works and Motivation}
Recognizing human activity in RGB-D videos is an active research area in computer vision and robotics \cite{mtap_survey}. Most of the approaches for depth based activity recognition are based on handcrafted features ranging from gradient, optical flow, HOG, interest points, etc. Oreifej \textit{et al.} \cite{c6} proposed Histogram of oriented 4D surface normals (HON4D), based on the assumption that, the normal orientation captures more information than the gradient orientation. The descriptors are acquired from 4D volumes for recognition. Spinello \textit{et al.} \cite{c9} used HOG to find the objects. Then Histogram of Oriented Depth value (HOD) is used for extracting depth information. Rahmani \textit{et al.} \cite{c5} proposed Histogram of Oriented Principal Components(HOPC) to tackle the issues related to viewpoint variations. Koch \textit{et al.} \cite{c12} proposed a volumetric approach for finding the depth estimates for all pixels and projected in the voxelized 3D space. Each depth estimate votes for a voxel probabilistically and the surfaces are extracted by thresholding. The handcrafted features are often found unable to find the motion in depth sequences, due to the 2D representation of the depth map, available with the data.

In order to overcome the difficulties of the handcrafted features in representing the depth motion efficiently, the concept of depth motion map was introduced \cite{c2,c7}. Depth map gives the position of the object whether it is near to the image plane or far from the image plane \cite{c2}. In \cite{c2}, Li \textit{et al.} presents a study on recognizing human actions from sequence of depth maps. The author used the concept of bag of 3D points. A small
number of 3d points are extracted to characterize the 3d shape of each salient posture of the performer and a Gaussian Mixture Model is used to effectively capture the statistical distribution of the points during the activity. Yang \textit{et al.} \cite{c7} proposed a method based on depth motion map (DMM) using HOG features. Chen \textit{et al.} proposed a depth feature representation technique based on an effective fusion of 2D and 3D auto-correlation of gradients features, capturing the shape and motion cues from the depth motion map \cite{chen_MTAP}. In DMM based methods, loss of spatial information between interest points is a concern for activity recognition.

A different school of thought relies on the skeleton information available from the kinect sensors, for recognizing the activity \cite{c11,c8}. Negin \textit{et al.} \cite{c11} proposed the concept of decision forest for features selection using skeleton joint position. Yang \textit{et al.} \cite{c8} used skeleton features from the RGB-D cameras. Pair wise difference between joints in a frame is calculated. Then the differences are checked with the preceding frame, and the difference is calculated and normalized using principle component analysis (PCA) and are classified using Naive-Bayes-Nearest neighbour method. Although the skeleton based approaches give high accuracy on certain datasets but are not applicable in videos with high level of occlusions.

Several other efforts have been made for depth based activity recognition\cite{c1,c4,BMVC}. Gonzalez \textit{et al.} \cite{c1} used a 3 stage model for gesture recognition which include background extraction, hand and face detection by extracting body skeletons and finally BICA architecture is used for activity recognition. Kong \textit{et al.} \cite{c4} merged RGB and depth features and learned a projection function based on both RGB and depth. A random forest classifier is applied on the DMM in \cite{BMVC}, where the geometry between scene layout and the human skeleton is considered for representation in the random forest. However, applying deep learned features for RGB-D activity recognition, still remains a less explored research  area. In \cite{ICCVW}, a comparative study between handcrafted and deep learned features is discussed, in the context of activity recognition.
\begin{figure*}
\centering
\includegraphics[width = \textwidth, height =7.5cm]{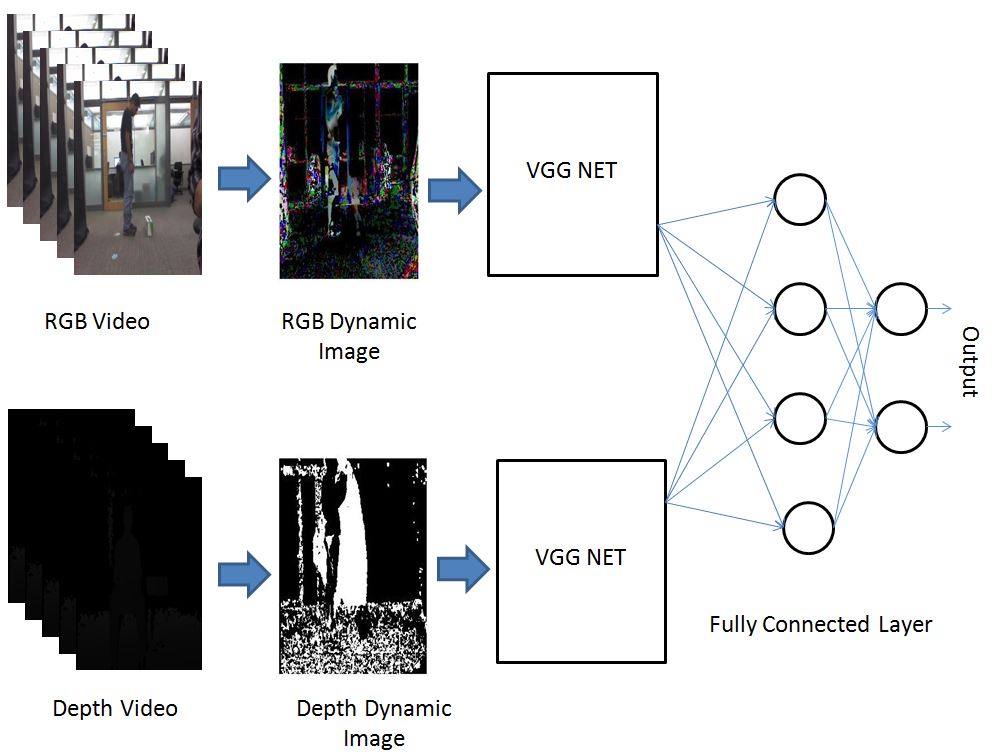}
\caption{A diagram showing the proposed approach for depth based activity recognition.}
\label{fig3}
\end{figure*}

A few techniques have been found in the literature based on deep learning based activity recognition in RGB-D videos \cite{THMS,AAAI}. A weighted hierarchical DMM is combined with a 3 layer CNN for feature extraction in \cite{THMS}. In \cite{THMS}, different view points are handled by rotating the 3D points of the DMM. In \cite{AAAI}, Wang \textit{et al.} proposed a deep learning based method where the same CNN is applied on both RGB and Depth frames, and the result feature is passed through a fully connected network, to get a good accuracy. Motivated by the success of Wang \textit{et al.} and Bilen \textit{et al.} \cite{c18}, we apply ResNext-101 on the dynamic images extracted from both RGB and depth videos, and pass the result vector through a fully connected layer. However, utilizing the dynamic images instead of using the whole video, reduce the computational overhead of the proposed method and provides a comparable accuracy with the state-of-the-arts.

\section{Proposed Method}
The proposed method has the following four steps: Dynamic images are extracted from the RGB and Depth videos separately; The dynamic images are processed using Gestalt philosophy of human perception; The processed dynamic images for RGB and Depth videos are fed into ResNext-101 \cite{resnext} separately; The extracted features from the two ResNext-101 networks are passed through a fully connected layer to obtain the final feature vector for classification. The overall method is shown in Figure \ref{fig3}. We discuss the steps sequentially.

\subsection{Extracting Dynamic Images}
The concept of dynamic images was first introduced by Fernando \textit{et al.} \cite{c15}. The procedure of \cite{c15} starts with ranking the consecutive frames $I_1,I_2,\ldots,I_t$ of the given video. Next, average of the features extracted from the frames over time, is computed. This means, average $A_t$ is obtained as follows:
\begin{equation}
A_t = \frac{1}{t}\sum_{\tau = 1}^{t} \psi(I_{\tau}),
\end{equation}
where $\psi(I_{\tau})$ represents feature vector extracted from each $I_t$.

For extracting the feature vectors we apply rank pooling directly from the frames, following \cite{c18}. In RGB-D videos, we apply rank pooling on RGB and Depth frames separately, to get two different dynamic images. The function $\psi(I_{\tau})$ stacks the RGB chanel components  of each pixel in $I_t$ to get a large vector. In case of Depth frame, this stacking is not necessary, as there is only a single chanel.

The ranking function for each frame assigns a score $S(t | d) = \langle d,A_t \rangle$ to the time $t$, where $d \in \mathbb{R}^{d}$ is a parameter vector. The parameters in $d$ is learned through a convex optimization problem using RankSVM \cite{ranksvm} as follows:
\begin{equation}
d^*=argmin_d{E(d)},
\end{equation}
where $d^*$ is the optimizing function of the vector $d$ and $E(d)$ is given by,
\begin{equation}
E(d) = \frac{\lambda}{2}\normalsize{d}^2+\frac{2}{T(T-1)}\sum_{p>t}{max\{0,1-S(p|d)+S(t|d)\}},
\end{equation}
where $T$ is the number of frames. In short, the above procedure transforms a sequence of $T$ frames into a single vector $d^*$, containing all the information of each frame of the video. This procedure is called rank pooling \cite{c15}. Note that, the video descriptor $d^*$ has the same number of elements as a single video frame and hence, can be considered as an
RGB image (in case of RGB video) or grayscale image (in case of depth video). The vectors $d_{rgb}^*$ and $d_{depth}^*$ obtained from the RGB and Depth videos respectively, are called dynamic images representing the respective videos.
\begin{figure*}
\centering
\includegraphics[width = \textwidth, height =7.5cm]{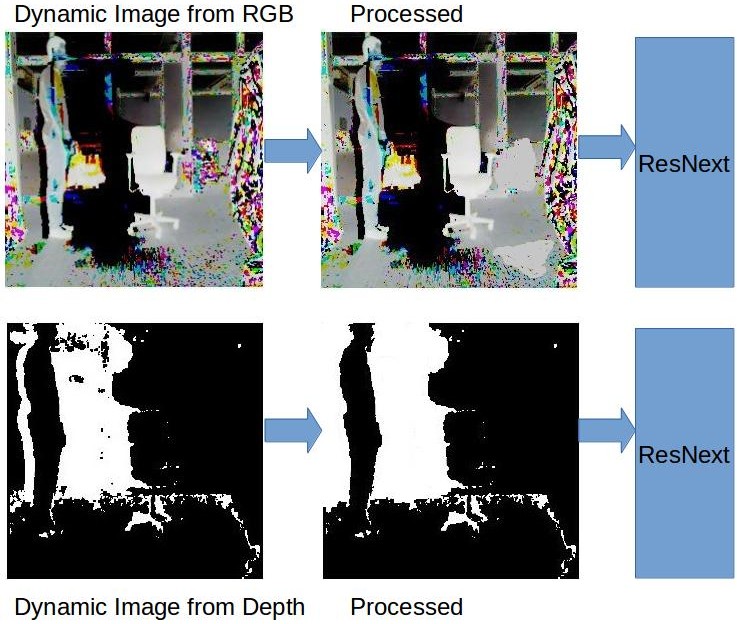}
\caption{An example of RGB and Depth dynamic image of a video from ``Pick up'' activity, and result of applying the proposed processing step on the dynamic images. Clearly, the Depth dynamic image is improved significantly, for segmenting the motion information, after the proposed processing step. Whereas, the RGB dynamic image is improved a little.}
\label{fig4}
\end{figure*}

\subsection{Gestalt philosophy for processing dynamic images}
We preprocess the dynamic images obtained from the RGB and Depth videos, before sending to a CNN classifier. The objective of the preprocessing step is to reduce the unwanted background motion information from the video, by eliminating sufficiently small connected components from the dynamic images. We assume that, the significant amount of motion information, during an activity, are concentrated only at the person and his/her surrounding area. Hence, the area describing the motion information in the dynamic image, concentrates around the performers' body. However, due to camera shaking, some motion information is seen at the background pixels as well. We believe that, the area describing the motion of background pixels are much smaller compared to the area describing the motion of the performers during the activity, as the camera movement can be considered as small. So, we eleminate the motion information from the dynamic images, where the area of the motion is significantly small. Figure \ref{fig4} shows an example dynamic image where the proposed preprocessing algorithm is applied.

As discussed above, area of the motion information in the dynamic images can be utilized to process the images. We obtain the connected components in the dynamic images and find the area of each of the components present in an image. We rank the normalized area of the connected components in sorted order. Next we find a suitable threshold on the normalized area of the components, to prune out the smaller components. We apply Gestalt theory of human perception \cite{mtap} to obtain a suitable threshold.

We first identify the connected components present in the dynamic images and calculate the area $a$. We multiply $a$ with the average intensity of motion in that area, to get the areal intensity $A_I$ of the connected component.
\begin{equation}
A_I=aI,
\end{equation}
where $I$ is the average intensity. The idea is that, background motion due to camera jerk and other atmospheric causes are generally of low intensities. Hence, the lower the value of $A_I$, greater is the possibility of the area to be eliminated in the processing stage. We rank the connected components according to their respective $A_I$ values and apply Gestalt philosophy \cite{mtap} to fix a threshold $\gamma$ on the $A_I$ value, so that, the components with $A_I$ value less than $\gamma$ is eleminated.

We vary the threshold $\gamma$ through all possible values of $A_I$, and for each value of $\gamma=\gamma^*$, we calculate the likeliness of $\gamma^*$ to be a suitable threshold for $\gamma$ using the Gestalt philosophy of human perception following \cite{mtap}. According to the Gestalt theory, a group of instances constitutes a random event in general. However, a group of instances can be considered as meaningful if the instances are unlikely to happen at random. For example, if most of the $A_I$ values are distributed over a specific region (these $A_I$ values can be considered as random), whereas, a few of them are significantly apart from the random values. These $A_I$ values having significantly different values compared to the random values are called ``meaningful'' values and hence, are pruned out from the set of all $A_I$ values. Consequently, the connected components corresponding to the pruned out $A_I$ values, are eliminated from the image.

For a given threshold $\gamma^*$, if $p$ is the prior probability that an arbitrary connected component has $A_I$ value greater than the threshold $\gamma^*$, then
\begin{equation}
p=1-\gamma^*,
\end{equation}
as $A_I$'s are normalized between 0 and 1. We assume that the $A_I$ values are independently and identically distributed (i.i.d.) with uniform distribution over the interval [0,1].

Let $n$ ($1 < n < N$, $N$ being the number of connected components in the image) be the minimum number of connected components with Gestalt quality (i.e., ``meaningful'' with respect to Gestalt theory) \cite{mtap}. Therefore, whether a particular threshold $\gamma^*$ is ``meaningful'' or not becomes a Bernoulli trial. Then the probability $P$ that the threshold $\gamma^*$ is meaningful is given by,
\begin{equation}
P=\sum_{i=n}^{N} {N \choose i} p^i (1-p)^{N-i}.
\end{equation}

As $P$ is associated with a Bernoulli trial, the distribution of $P$ follows a Binomial distribution and hence, the expected number of occurrences of the ``meaningful'' event (i.e., the $A_I$ value of the connected component is significantly small), called the number of false allarms (NFA), is given by,
\begin{equation}
NFA=\delta P,
\end{equation}
where $\delta$ is the number of $\gamma$ values in [0,1], tried to find the optimum $\gamma$. If $NFA$ for a connected component is less than a predefined number $\epsilon$, then the corresponding threshold value $\gamma^*$ is called $\epsilon$-meaningful and the connected component is pruned out from the image. We set $\epsilon=1$ as in \cite{mtap}.

\subsection{Classification Using Resnext}
We feed the dynamic images obtained from the RGB and Depth videos to two different CNNs. We concatenate the feature vectors obtained from the CNNs for RGB and the Depth videos, and feed into a fully connected layer for final classification. We tried with two different CNNs: VGG16 \cite{c17} and Resnext-101 \cite{resnext} following \cite{c18}, where Resnext-101 gives better result, as in \cite{c18}.

\section{Experiments \& Datasets}
The execution overhead of the proposed convolution is less as it uses a simple 3x3 convolution matrix, both for Resnext-101 and VGG16. We experiment with one very deep CNN (Resnext-101) and one shallow CNN (VGG16), in order to have a clear perception over the proposed dataset. Although the proposed dataset is not a huge dataset, still Resnext-101 produced better accuracy compared to VGG16. The experiments are performed on two datasets: MSR Action 3D dataset \cite{data_msr} and the proposed dataset. Both the datasets have been divided into two parts with 70\% for training (randomly chosen) and rest for testing. The two datasets are described as follows.

\subsection{MSR Action 3D Dataset}
The MSR Action 3D dataset conssts of 20 action classes performed by 10 subjects, where each subject performs each action 2 or 3 times. The total number of videos is 567 with resolution 640x240. The data was recorded using a depth sensor like Kinect device.

\subsection{Proposed Dataset}
As discussed in the Introduction, the existing datasets do not contain multiple performers in a single video. Hence, we introduce a challenging RGB-D dataset with 7 action classes. The actions are pick \& throw (PT), general loitering (GL), playing badminton (BD), punching (PN), pointing (PO), discussion (DS) and high throw (HT). The dataset was collected using Microsoft Kinect V2 sensor to get both RGB and depth videos. We have QHD resolution to capture the actions. Our dataset have the following challenges:
\begin{enumerate}
\item Multiple persons performing the same action.
\item Two different actions performed simultaniously.
\item Camera jitter and motion blur often make the dataset more challenging, as the videos were collected from hand held sensor.
\item Illumination changes are present, e.g., light will be dimming in some parts of the video.
\item Cluttered and uneven background in each video.
\end{enumerate}
The proposed dataset contains 73 RGB and Depth video sequences for the 7 action classes.

\section{Results}
The proposed method shows better results compared to the state-of-the-art methods, on both the datasets. Table \ref{tab1} shows the efficacy of the proposed method compared to the competing methods. For the MSR Action 3D dataset \cite{data_msr}, the accuracy of each of the competing methods is above 90\%, whereas, for the proposed dataset, the accuracy of each of the methods are very low, due to the challenges involved in the dataset. Also we can observe from Table \ref{tab1} that, the proposed preprocessing step plays a vital role in recognition. Minimization of the motion of background pixels due to camera jerk during activity, helps the proposed method to achieve a better accuracy. Clearly, Resnext performs much better than VGG16 on both the datasets.
\begin{table}[]
\caption{Performance of the proposed approach compared to the state-of-the-art, applied on the MSR Action 3D \cite{data_msr} and the proposed datasets.}
\begin{tabular}{|l|l|l|}
\hline
\multirow{2}{*}{Methods} & \multicolumn{2}{l|}{Accuracy (\%)} \\ \cline{2-3} 
                         & MSR Action 3D  & Proposed  \\ \hline
Chen \textit{et al.} \cite{chen_MTAP} &      94.87    &       45.24       \\ \hline
Kong \textit{et al.} \cite{c4}       &       91.18    &       42.22       \\ \hline
Baek \textit{et al.} \cite{BMVC}     &       95.24    &       49.94       \\ \hline
Wang \textit{et al.} \cite{THMS}     &       94.92    &       48.44    \\ \hline
Wang \textit{et al.} \cite{AAAI}     &       95.37    &       51.23    \\ \hline
Proposed method using VGG16     	 &       94.44    &       44.44    \\ \hline
Resnext without Proposed processing  &       94.73    &       50.12       \\ \hline
Proposed method with processing	     &       96.17    &       54.94       \\ \hline
\end{tabular}
\label{tab1}
\end{table}

The proposed dynamic image based method for activity recognition, eliminates the effect of background clutter by concentrating only on the motion of the moving pixels and preserving the motion information of the whole video in a single image. This dynamic image preserves the temporal information during activity, because of the motion information only at the moving pixels. The dynamic image is also capable of preserving the necessary spatial information due to the motion location during activity.
\begin{table}[]
\caption{Confusion matrix for the proposed method applied on the proposed dataset.}
\begin{tabular}{|l|l|l|l|l|l|l|l|}
\hline
Activities & PT & GL & BD & PN & PO & DS & HT \\ \hline
PT & 5 & 0 & 2 & 1 & 0 & 0 & 2 \\ \hline
GL & 0 & 7 & 0 & 1 & 0 & 3 & 0 \\ \hline
BD & 0 & 0 & 6 & 1 & 0 & 0 & 4 \\ \hline
PN & 0 & 0 & 1 & 5 & 3 & 0 & 1 \\ \hline
PO & 0 & 0 & 1 & 4 & 5 & 0 & 0 \\ \hline
DS & 0 & 2 & 0 & 1 & 0 & 7 & 0 \\ \hline
HT & 1 & 0 & 4 & 1 & 0 & 0 & 5 \\ \hline
\end{tabular}
\label{tab2}
\end{table}

Table \ref{tab2} shows the confusion matrix for the proposed method on the proposed dataset. Table \ref{tab2} depicts the level of challenge involved in the proposed dataset, where several activities are confusing for all the state-of-the-art methods, due to the complicacy of the dataset with activities of similar appearances. For example, the activities such as, Punching \& Pointing, High throw \& Badminton, Discussion \& General loitering confuse the existing methofds due to their similar appearances.

As discussed, the proposed approach generates a single image from a video, which is enough for the action classification. This idea reduces the memory overhead during execution, and provide much faster execution compared to the state-of-the-art techniques. The results of all the experiments reported in this paper, have been generated within a few hours for the MSR Action 3D dataset. Such fast execution is uncommon in any video processing method based on deep architectures.

\section{CONCLUSION}
We have proposed a novel technique for recognizing human activities in RGB-D videos, based on dynamic images generated from the videos. The proposed method is validated with a benchmark dataset and a new challenging dataset introduced in this paper. We have adopted the recent technique of human activity recognition using dynamic images, and extended the concept to recognize activities in RGB-D videos. The data size is a big concern for any RGB-D video analysis techniques, whereas, the proposed dynamic image based approach can provide an efficient yet space- and time-effective procedure for any RGB-D video analysis tasks. Moreover, the processing step over the dynamic images, to reduce the unwanted background motion, provides an edge for the proposed approach over the competing methods. The proposed processing step based on Gestalt theory can be effective for several vision related application areas including semantic segmentation, tracking, noise removal.

\end{document}